\documentclass[letterpaper, 10 pt, conference]{ieeeconf}  

\IEEEoverridecommandlockouts                              

\usepackage{graphics} 
\usepackage{epsfig} 
\usepackage{mathptmx} 
\usepackage{times} 
\usepackage{amsmath} 
\usepackage{amssymb}  
\usepackage{tabu}
\usepackage{booktabs}
\usepackage{cite}
\usepackage{xcolor}
\usepackage[colorlinks=true,linkcolor=blue, urlcolor=black, citecolor=blue]{hyperref}
\usepackage[caption=false,font=normalsize,labelfont=sf,textfont=sf]{subfig}

\title{\LARGE \bf
MorphoNavi: Aerial-Ground Robot Navigation with Object Oriented Mapping in Digital Twin
}

\author{Sausar Karaf*, Mikhail Martynov*, Oleg Sautenkov, Zhanibek Darush, Dzmitry Tsetserukou
\thanks{*Equal contribution}
\thanks{Authors are with Intelligent Space Robotics Laboratory, CDE, Skoltech,
         Bolshoy Boulevard, 30, bld. 1, Moscow 121205, Russia
        {\tt\small {sausar.karaf, mikhail.martynov, oleg.sautenkov, zhanibek.darush, d.tsetserukou} @skoltech.ru}}%
}
\begin{document}

\maketitle
\thispagestyle{empty}
\pagestyle{empty}

\begin{abstract}
This paper presents a novel mapping approach for a universal aerial-ground robotic system utilizing a single monocular camera. The proposed system is capable of detecting a diverse range of objects and estimating their positions without requiring fine-tuning for specific environments. The system's performance was evaluated through a simulated search-and-rescue scenario, where the MorphoGear robot successfully located a robotic dog while an operator monitored the process. This work contributes to the development of intelligent, multimodal robotic systems capable of operating in unstructured environments.
\end{abstract}

\section{Introduction}

Robotics has experienced rapid advancements in recent years, with Vision-Language Models (VLMs) emerging as a powerful tool for mission execution based on RGB images. Since VLMs require only an image and a text prompt as input, they eliminate the need for expensive and specialized sensors such as LiDARs and depth cameras. This simplicity and cost-effectiveness suggest that vision-language-based control will play a crucial role in the future of robotics, with cameras becoming the primary sensor for most robotic systems.

In this paper, we introduce a novel mapping approach designed for a universal air-ground robotic system using a single monocular camera. This method facilitates mission planning and path execution using built-in path-following techniques while enabling seamless integration of VLM modules without additional training. Unlike conventional mapping techniques, our approach minimizes the need for high-bandwidth communication channels between the robot and the control interface, making it more efficient and scalable.

Additionally, our method retains significantly more information relevant for high-level reasoning and downstream neural network processing. Traditional mapping approaches, such as obtaining point clouds, octomap~\cite{OctoMap}, and mesh restoration techniques, primarily focus on preserving object shapes. In contrast, our approach also preserves the semantic meaning of objects, enabling higher-level understanding, such as inferring room functions, planning multi-stage missions, and rapidly scanning environments without requiring time-consuming detailed investigation of environment. While this comes at the cost of reduced map accuracy, our experimental results (Section~\ref{experiments}) demonstrate that the achieved precision is sufficient for global robotic navigation. This method can serve as both an augmentation to existing robotic perception systems—enhancing their interaction with large language models—and as a cost-effective alternative for robots equipped with only cameras and limited computational resources.

\begin{figure}[t!]
\centering
\includegraphics[width=1\columnwidth]{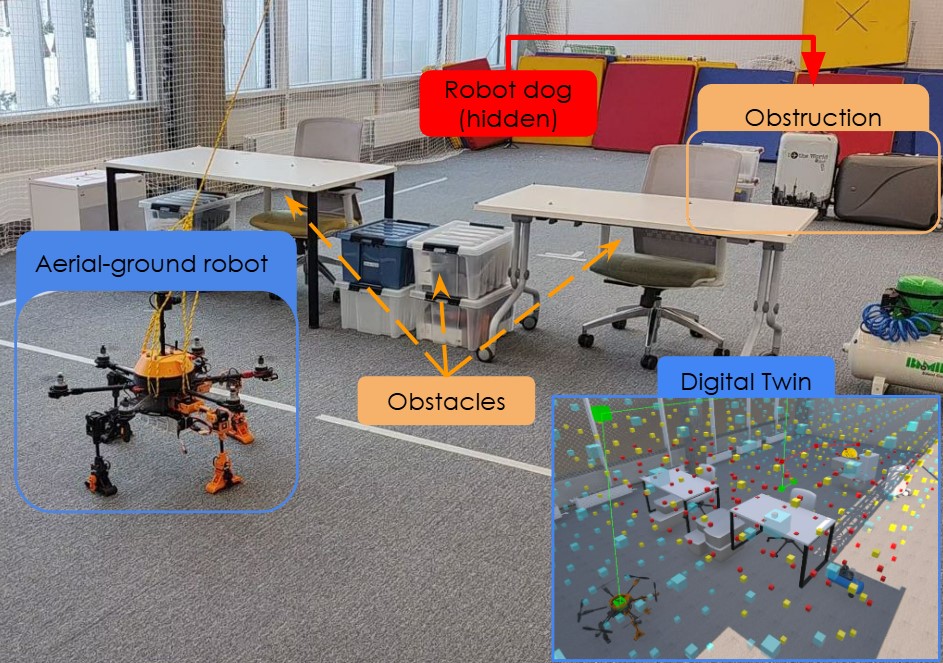}
\caption{Experimental setup. The mission is to overcome obstacles and search for the hidden robot.}
\label{fig:Experiment_part1}
\end{figure}

Our approach addresses several key tasks: object recognition, image segmentation, depth completion, and distance estimation. Object recognition is aimed at identifying the maximum number of objects and, if possible, their properties, such as the composition of the material. The segmentation stage consists in highlighting the identified objects in the image, from which we can conclude about the size of the objects and use this in the next stage. Depth mapping and distance estimation are interrelated. The main purpose of depth estimation is to preserve the relative distances between objects. By knowing the characteristic length or width or height of at least one object, the entire depth map can be scaled accordingly. Many common objects, such as cups, tables, vehicles, mature trees, and even people, most often have standard dimensions with minor deviations. Such reference objects will almost certainly occur in a randomly selected environment. The deviation in the size of the reference objects affects the accuracy of the entire method, but the remoteness of the object eliminates this error.

Our methodology is founded on the premise that exact color representation and shape replication are less critical for environmental understanding than semantic meaning. By recognizing an object’s approximate dimensions, purpose, and contextual role, we can infer its past and future interactions within the environment—akin to how humans comprehend objects from minimal information such as name of object. For example, seeing a person walking towards a taxi, you can predict that the car will soon move and you don't need to walk in front of it. In this case, the accuracy of the object's shape and location is not as significant as understanding that the route immediately needs to be rebuilt behind the car. This ability to derive meaning from sparse data underscores the potential of our approach for robust and efficient robotic navigation and decision-making.

\section{Research Background}

Monocular depth estimation has emerged as a critical component for robotic perception, enabling the reconstruction of 3D spatial information from a single RGB image. \textit{ZoeDepth}~\cite{zoedepth} represents a significant advancement in this domain, combining relative depth estimation (RDE) with metric depth estimation (MDE) through a two-stage training process: pre-training on relative depth datasets followed by fine-tuning on metric datasets like NYU-Depth V2. This hybrid approach achieves high accuracy in preserving relative distances between objects. Similarly, \textit{Depth-Anything}~\cite{depthanything} introduces a zero-shot, backbone-scalable framework that leverages large-scale unlabeled data to produce robust depth predictions across diverse environments.

The \textit{YOLO} (You Only Look Once) family of models~\cite{yolo}, known for their real-time performance, excels in detecting predefined classes with high speed and efficiency. However, their limitation to a small number of classes requires additional training for broader applicability, which may not align with our goal of minimal retraining. Similarly, \textit{Detectron2}~\cite{detectron2} supports object detection and instance segmentation but shares YOLO’s dependency on supervised training for specific datasets. In contrast, zero-shot and open-vocabulary detectors like \textit{Grounding DINO 1.5 Pro}~\cite{groundingdino15pro} and \textit{DINO X}~\cite{dinox} offer transformative potential by integrating vision and language to recognize objects beyond pre-trained categories. \textit{OWL-ViT}~\cite{owlvit} and its successor \textit{OWLV2}~\cite{owl2} further enhance this capability with transformer-based architectures, making OWLV2 a stronger candidate for precise object recognition in our framework.

Image segmentation complements object detection by delineating object boundaries, providing critical size and shape information for depth scaling and path planning. The \textit{Segment Anything Model (SAM)}~\cite{sam} introduces a promptable segmentation system capable of zero-shot generalization to unseen objects, aligning with rapid environmental scanning without exhaustive scene reconstruction. Its successor, \textit{Segment Anything Model v2}~\cite{samv2}, builds on this foundation with enhanced scalability and performance, offering a robust solution for segmenting objects in dynamic robotic contexts. These models enable our approach to highlight identified objects and infer their spatial extents, feeding into subsequent depth estimation stages.

The advent of zero-shot, backbone-scalable models like \textit{DINO 1.5}~\cite{dino15} and \textit{DINO X}~\cite{dinox} reflects a shift toward flexible, training-free perception systems. DINO 1.5  employs a transformer-based encoder-decoder architecture to achieve open-world object understanding. These models prioritize high-level reasoning—such as inferring object purpose or contextual role—over exact shape replication, mirroring human-like sparse-data comprehension. 

Traditional mapping techniques, such as point clouds from SLAM or mesh-based reconstructions, excel in geometric fidelity but often discard semantic context, requiring high computational resources and bandwidth for processing and transmission. In contrast, models like ZoeDepth, Depth-Anything, OWLV2, and SAM to retain both spatial and semantic information with minimal overhead. While YOLO and Detectron2 offer speed and precision for known classes, their need for retraining limits scalability in open environments. Zero-shot models (e.g., Grounding DINO 1.5 Pro, DINO X) and scalable frameworks (e.g., SAM v2, Depth-Anything) bridge this gap, providing the adaptability and efficiency.


Vision-Language Models (VLMs) such as Molmo~\cite{Molmo} and ChatGPT~\cite{ChatGPT} have revolutionized the integration of visual and textual data, enabling tasks like zero-shot object recognition and image captioning by aligning 2D image features with natural language descriptions. However, these models are not inherently designed to operate in 3D space, as their training focuses on 2D image-text correspondence rather than spatial reasoning or depth perception. This limitation poses challenges for robotic applications requiring 3D environmental understanding, such as navigation and obstacle avoidance. To address this, Vision-Language-Action (VLA) models, such as RT-1~\cite{rt1} and PaLM-E~\cite{palme}, have been developed to combine vision, language, and action planning. These VLA models demonstrate impressive capabilities in robotic manipulation and navigation but rely on extensive, task-specific datasets—e.g., robotic trajectories or simulated 3D scenes—which are resource-intensive to collect and often tailored to narrow domains. 

Since aerial-ground robots (AGRs) are still an emerging field, there is limited research focused on path planning for these systems. Most existing path planners are developed for heterogeneous robotic systems. However, HE-Nav~\cite{HENAV} provides path planner building upon completed map in cluttered environments, and OMEGA~\cite{OMEGA} presented AGR-Planner utilizes local map and employs kinodynamic A* search and energy-efficient planning. These algorithms can be used in AGVs~\cite{pathplan},  they require careful selection of safe take-off and landing points for aerial operations.

\section{System Overview}

Our system consisted of three main parts, a robot, a laptop with a control interface, and an environment with a localization system. All calculations were performed either on the robot (for control) or on a personal computer (for mapping). We used the Unity Game Engine, because in the future it will allow us to simulate the movement of objects and predict their behavior virtually, without processing real data.

\subsection{MorphoGear: Aerial-Ground Robot} 
\begin{figure}[htb!]
\centering
\includegraphics[width=0.8\linewidth]{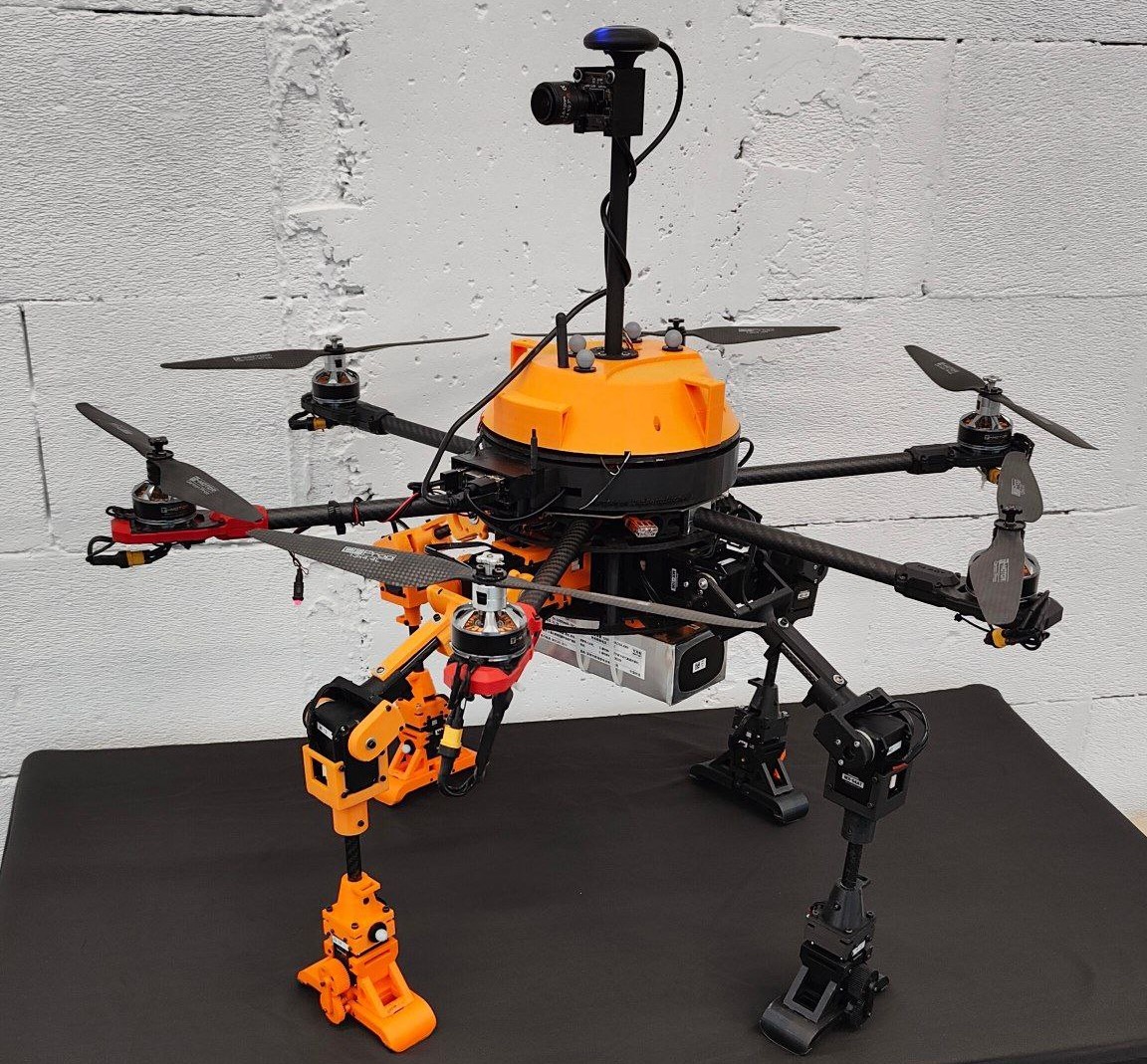}
\caption{Aerial-Ground Vehicle MorphoGear.}
\label{morphogear}
\end{figure}

 MorphoGear (Fig.~\ref{morphogear}) - unmanned aerial ground vehicle (AGV) with morphogenetic gear for terrestrial locomotion, object grasping, and aerial motion. We used this robot because it implements a unique interaction with the environment. It can overcome obstacles that are inpathable to ground robots. It can stop on the ground to take high-quality data, as well as for the time of processing and waiting for new commands, ensuring long-term operation, which is impossible for drones. In addition, it allows interaction with objects through manipulation, which will be implemented in future work using our control interface. 

The robot includes the following hardware: companion computer OrangePi 5b, flight controller OrangeCube, custom limb controller based on STM32, camera ELP-USBFHD05H 2MP 2.8-12mm 1:1.4 1/2.7" MJPEG. Software: Robot Operating System (ROS2 Iron), which contains Python nodes for high-level commands and mavros. ROS\# generates limbs motion as a continuation of previous work~\cite{MorphoGear_AIM}, moving the scripts from Unity to internal robot calculation (used as asynchronous methods). The flight controller is flashed with Ardupilot v4.4.1. Software running on Ubuntu Server - Jammy Jellyfish (22.04)

\subsection{GUI: Ground Station}
The operator place is equipped with a ground station - a laptop with pre-installed Unity (2022.3.9) and Python (3.10).
Laptop: Intel i7-1165G7, 16GB RAM, MX450 laptop 2Gb VRAM.
We have developed a simulator for this robot. It is a digital twin of the robot, allowing for initial experiments to be carried out virtually. It is also used as a control panel for the robot. In this case, the robot sends its state to Unity, and it acts only as a visualization. Since all robot control works on ROS2 Iron we use ROS-TCP-Connector (v0.7.0) by Unity-Technologies to send commands from Unity to the robot.

\subsubsection {Unity scene initial setup}
Initially, we built an empty room in which we conduct experiments. A digital twin of the robot was created, with bidirectional communication, commands in one direction, the state in the other.
Figure~\ref{interface} shows an example of the operator's view. The operator can move freely around the scene.

\subsubsection {Path planner}

We developed the Aerial-Ground A* algorithm to perform path planning. It is a modification of the algorithm that was presented in work~\cite{MorphoMove_SMC}. In Fig.~\ref{interface} red cubes mean occupied nodes (inpathable), yellow cubes mean dangerous nodes surrounding occupied (significantly increased cost), blue nodes are free grid nodes. Ground layer has no cost (walking is a safe type of motion), the first layer has high cost (which means that takeoff and landing are the most unsafe operations), and the rest layers have normal cost (flight is less preferable than walking indoors). 
The interface of setting the path has green lines that mean the registered path, and red lines that mean the calculated path to be registered. If the path is registered, next calculation starts from the last registered node. This allows designing a complex path with multiple stops.
After path planning, it is converted to a mission. The path splits to ground motion, takeoff spot, flight, land, ground motion again, and etc. All those parts of the mission require switching between ROS nodes and packages, as some of them may have conflicts with each other. We developed a mission manager that switches nodes according to the type of motion and part of the path. It initializes nodes and reboots hardware if it is needed. 

\subsubsection {Object recognition script}
A python script has been prepared to generate the map. The script receives one photo as input and returns the coordinates of all recognized objects. The script is adapted to quickly change neural network models. The script solves the following tasks: object recognition, image segmentation, depth completion, distance estimation. As output, here are object names and their positions.

\subsubsection {Object spawner}
After recognizing objects using our Python script, a JSON file is created containing information about the object and its position relative to the robot. This file is placed in Streaming Assets in Unity, which allows files to be loaded into the simulation on the fly without the need to reboot and recompile the scene. The script takes data from JSON and instantiates objects with respect to their name in the local coordinate system of the robot, then reassigns them to map with the global frame. Unity has a list of models that can be used. The models may not match the real objects, but they retain the meaning of the object and its approximate dimensions. In addition, this approach significantly speeds up scene rendering and the amount of data transferred. 

\begin{figure}[htb!]
\centering
\includegraphics[width=1\linewidth]{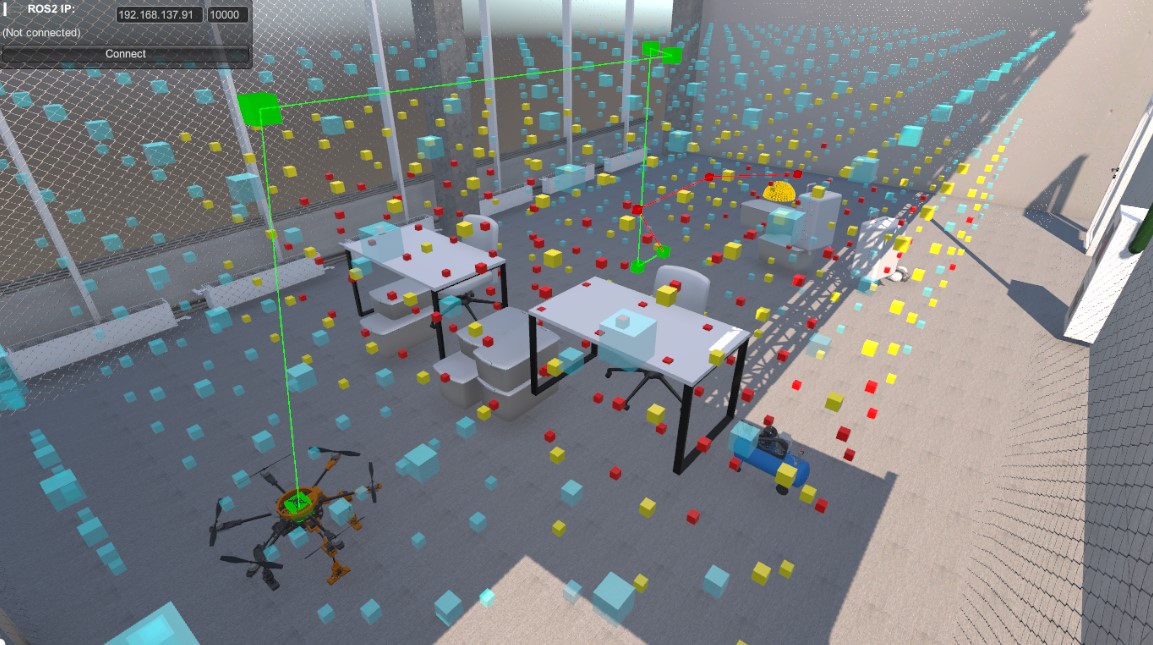}
\caption{Virtual simulation and visualization for MorphoGear.}
\label{interface}
\end{figure}

\subsection{Environment}
The experiments were conducted in a room measuring 6x10x4 meters. The workspace is limited by a net; for safety purposes, the path planning grid was 5x8x3 meters. There were objects: two different desks, two office chairs, two different suitcases, a compressor, an office cabinet, boxes, and sports mats (did not participate in the experiment, but were previously used to debug algorithms because they have an ideal square meter size). The room is equipped with a localization system, VICON.

\section{Algorithm of Mapping} \label{mapping_alg}

Effective navigation of a ground-aerial robotic system requires a map that accurately represents key environmental elements. The system must detect objects of interest and estimate their positions based on their known geometric dimensions. Once the 3D objects are positioned within the scene, the robot generates and follows a path to a user-defined destination.

\begin{figure}[tb]
  \centering
  \includegraphics[width=1\columnwidth]{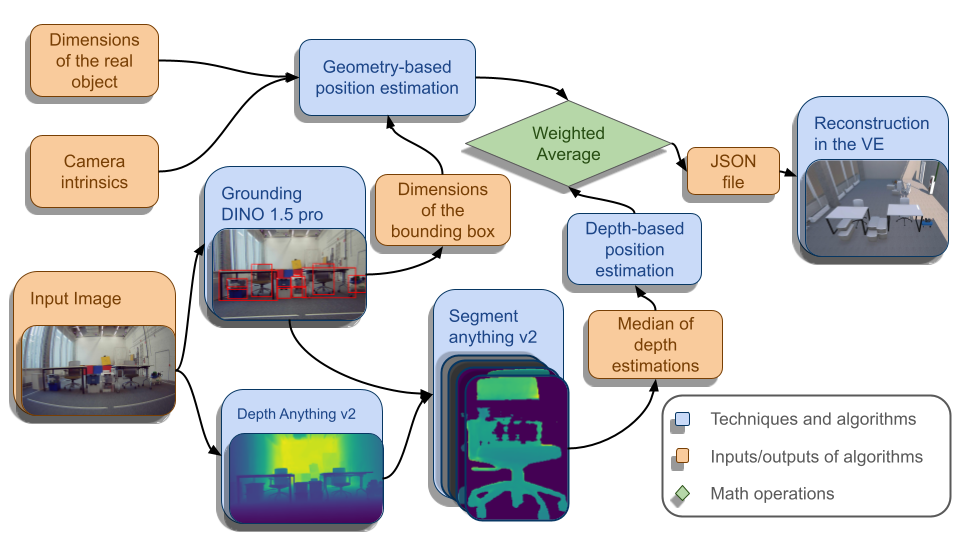}
  \caption{The system architecture of the mapping pipeline.}
  \label{fig:mapping architecture}
\end{figure}

The proposed system (Fig.~\ref{fig:mapping architecture}) operates using a single monocular RGB image as input. During development, multiple object detection models, including OWLv2, OWL-ViT, and DINO-X, were evaluated. Among these, OWLv2 and Grounding DINO 1.5 Pro demonstrated the best performance in our testing environment (Section~\ref{experiments}) and were selected for implementation. Given the known object dimensions, camera intrinsics, and the bounding box obtained from the object detector, the object’s distance is estimated using the following formula:

\begin{equation} 
\label{eq:sdf_cost} 
d = f \frac{h_m}{h_{px}}, 
\end{equation}
where $f$ is the focal length, $h_m$ is the object's real-world height (in meters), and $h_{px}$ is its height in pixels.

To refine this estimate, we leverage Depth Anything v2 in combination with Segment Anything v2 to obtain a depth estimate using a deep learning-based approach. A segmentation mask, generated by Segment Anything, is applied to the depth map, and the median depth value within the masked region is extracted. The final object distance is computed as a weighted average of the two estimates, with 80\% weight assigned to the geometry-based estimation and 20\% to the depth-based estimation. It is worth noting, that not all object dimensions were known. In the case of unknown object dimensions only depth estimations were used. The processed object data is then packaged into a JSON file and transmitted to a Unity-based simulation environment.

\section {Experiment} \label{experiments}

To evaluate the proposed system, we conducted a simulated search and rescue scenario. We have come up with a real case where a robot dog has encountered a problem and requires outside intervention. The test environment included obstacles such as tables, boxes, and chairs, restricting the robot's initial field of view and making full-map observation challenging. A Unitree Go1 robotic dog was placed behind these obstacles, and the MorphoGear robot's objective was to locate it using the proposed system. Throughout the experiment, an operator monitored the reconstruction process and verified that the robot functioned as expected.

The system's performance was assessed based on three key metrics: the ratio of detected objects to the total number of target objects, the accuracy of object detection, and the calculation time.

\subsection{Choice of the Detection Model}

To optimize the navigation of the aerial-ground robot within a complex environment, selecting the most suitable object detection model is essential. We evaluated several state-of-the-art models in the context of our experimental setup, aiming to gather data on detection ratio and computational time. The results of these evaluations are summarized in Table~\ref{tab:comp_detect}.

\begin{figure}[tb]
\centering
\includegraphics[width=0.9\columnwidth]{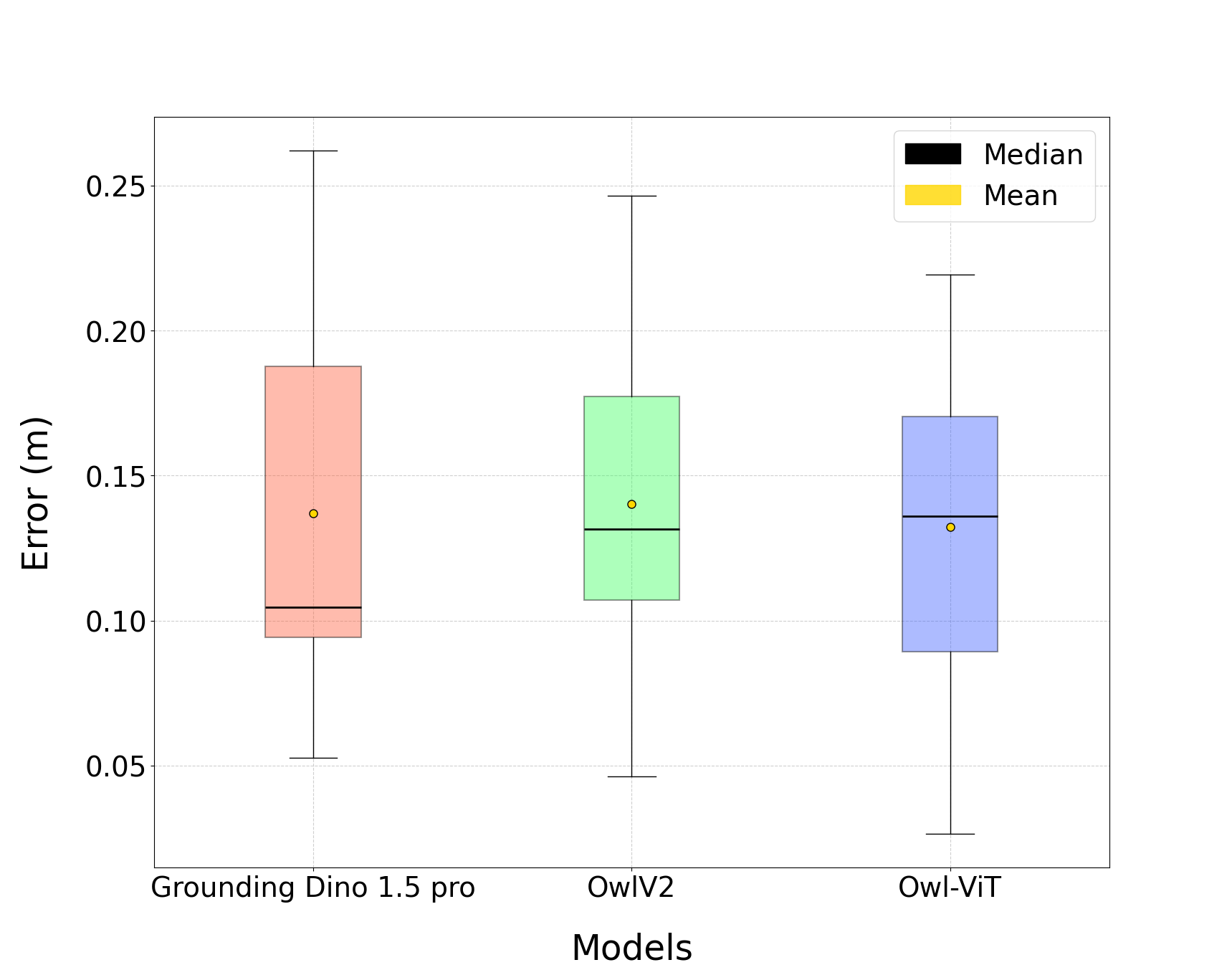}
\caption{Accuracy of position estimates.}
\label{fig:detection_accuracy}
\end{figure}

Inference with the OWLv2 and OWL-ViT models was performed locally on the laptop, while the Dino-X, Grounding DINO 1.6 Pro, and Grounding DINO 1.5 Pro models were accessed through their respective APIs. The evaluation revealed that, while OWL-ViT achieved the fastest computation time, its detection accuracy was lower compared to the other models. In contrast, the Grounding DINO 1.5 Pro model exhibited the most balanced performance, achieving a high detection ratio with moderate computation time.
\begin{table}[h]
  \caption{%
        Comparison of Detection Models.%
  }
  \label{tab:comp_detect}
  \scriptsize%
  \centering%
  \begin{tabu}{%
          X[0.5,l]%
                *{2}{X[0.25,c]}%
        }
        \toprule
        Model & Detection Ratio (\%) & Calculation Time (sec)\\
        \midrule
        Dino-X                      & 90.4        & 7.26 \\
        Grounding Dino 1.6 pro      & 93.2        & 7.07 \\
        Grounding Dino 1.5 pro      & 97.4        & 7.34 \\
        Owl v2                      & 69.5        & 13.21 \\
        Owl-ViT                     & 37.5        & 4.77 \\
        \bottomrule
  \end{tabu}%
  \vspace{-0.3cm}
\end{table}
Further analysis of pose estimation accuracy, as shown in Figure~\ref{fig:detection_accuracy}, indicated similar trends, with the Grounding DINO 1.5 Pro slightly outperforming the other models. This is particularly significant, as the quality of pose estimation is directly dependent on the accuracy of the bounding box predictions. Inaccurate bounding boxes, such as those failing to fully encompass the detected object, can lead to erroneous pose estimates.
Based on these findings, the Grounding DINO 1.5 Pro model demonstrated the best overall performance and was chosen for the subsequent experiments. This model strikes an optimal balance between detection accuracy and computational efficiency, making it the most suitable choice for our system.

\subsection{Experimental Setup}
A specific test case was designed to evaluate the system's capabilities and limitations (Fig.~\ref{fig:Experiment_part1}). In this setup, obstacles were arranged in a straight line in front of the starting position, dividing the search area into two separate regions. The robotic dog was positioned behind an obstruction made of stacked boxes.


The robot's task was to detect nearby obstacles, plan a path to navigate past them, and, upon recognizing the obstruction, move behind it to locate the robotic dog. This arrangement of the obstacles was chosen to, as well, validate the MorphoGear robot’s ability to transition between terrestrial and aerial locomotion.

\subsection{Experimental Procedure}

At the start of the mission, the aerial-ground vehicle captured an initial image of the environment. This image was processed by the mapping pipeline, which computed object positions and sent them to the Unity-based GUI for visualization and planning. Using the generated map and the robot's position, an obstacle grid was constructed and used by an A* algorithm to plan a trajectory for the MorphoGear robot. The path-planning results are shown in Fig.~\ref{interface}. 

\begin{figure}[tb]
\centering
\includegraphics[width=1\columnwidth]{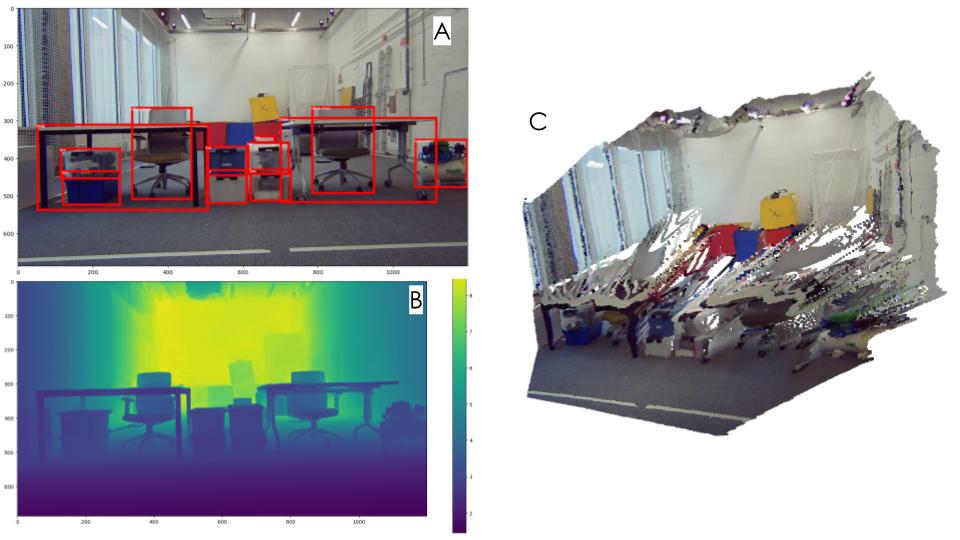}
\caption{Experimental results: (A) Object recognition, (B) Depth completion, (C) Point cloud.}
\label{fig:mapping pipeline}
\end{figure}

Figure~\ref{fig:mapping pipeline} illustrates the mapping pipeline described in Section~\ref{mapping_alg}. As shown in Fig.~\ref{fig:mapping pipeline} (c), the generated point cloud exhibits noise at object edges, hindering safe landing in occluded areas. Additionally, occluded objects are distorted, leading to inaccurate pose estimations. To improve this, our pipeline integrates both depth and object size information, resulting in more accurate pose estimates and safer navigation.


\subsection{Experiment Results}

The system's performance was evaluated based on the two predefined metrics: object detection ratio and accuracy of position estimates.

\subsubsection{Object Detection Ratio} The system successfully detected and localized 97.4\% of the target objects present in the scene. The performance varied depending on factors such as object occlusion and lighting conditions. Objects positioned behind dense obstacles were more challenging to detect, leading to failed detection in certain cases.

\subsubsection{Position Estimation Accuracy} The position estimates generated by the proposed system were compared to the ground truth position from the VICON position system. The average position error of the system is 13.6 cm, demonstrating reliability for our application's localization of detected objects. Errors primarily stemmed from partial occlusions. A graph of detection accuracy is shown in Fig.~\ref{fig:detection_accuracy}.

\subsubsection{Computation Time} 

The average time required for scene reconstruction was measured to assess system efficiency. The mapping system processed each image in 7.3 seconds on average, including object detection and obstacle mapping. These results indicate that the system operates within a feasible time frame for real-time search and rescue applications, though further optimization could enhance responsiveness.


The results indicate that the MorphoGear robot is able to successfully navigate through the given environment while identifying obstacles to locate the robotic dog.

\subsection{Discussion and Limitations}
While the proposed system demonstrated robust performance in the experimental scenario, several limitations were observed:

\subsubsection{Occlusion Challenges} The position accuracy of the objects is very dependent on the bounding box dimensions, thus dense clutter sometimes obstructs the field of view of the camera, leading to incorrect object positioning.

\subsubsection{Unknown shapes and varying orientations} The monocular-based distance estimation method performed well for known object sizes but exhibited inaccuracies for objects with irregular shapes or varying orientations.

\subsubsection{Computation Time} Real-time processing was not achieved in this version of the system. Optimizations in model inference speed could enhance system responsiveness and will be considered.

\section{Conclusion and Future Work}
In this paper, we presented a mapping approach for a universal aerial-ground robotic system utilizing a single monocular camera. The proposed system demonstrated the ability to detect a diverse range of objects and estimate their positions without requiring fine-tuning for specific environments. Experimental validation was conducted through a simulated search-and-rescue scenario, where the MorphoGear robot successfully located a robotic dog while an operator monitored the process. The system achieved an object detection rate of 97.4\%, with an average position estimation error of 13.6 cm and an average processing time of 7.34 sec per image.

While the system performed well in controlled laboratory conditions, several areas for improvement remain. Incorporating orientation estimation into the pipeline will lead to more accurate position calculations, particularly in cluttered environments. Additionally, occlusion remains a challenge, as partially visible or obstructed objects often result in incorrect position estimates. Future work will explore hierarchical and deep learning-based approaches to mitigate these issues.

Another crucial direction for future research is integrating the proposed mapping system with vision-language models (VLMs). By providing VLMs with structured scene information from our mapping pipeline in addition to the raw monocular image, we aim to enhance their spatial understanding and cognitive reasoning capabilities. This integration is expected to significantly improve the system’s ability to interpret complex environments, leading to better decision-making and task execution in real-world applications. Furthermore, we will investigate real-time optimization strategies to reduce processing latency, making the system more responsive for dynamic search-and-rescue scenarios.

Ultimately, this work contributes to the development of intelligent, multi-modal robotic systems capable of operating in unstructured environments. By addressing the identified limitations and expanding the system’s capabilities, we move closer to achieving robust, autonomous aerial-ground navigation and perception for real-world deployment.

\addtolength{\textheight}{-12cm}   






\bibliographystyle{IEEEtran}
\bibliography{reference}

\end{document}